%% file: main.tex
\documentclass[10pt,twocolumn,letterpaper]{article}

\usepackage{iccv}
\usepackage{times}
\usepackage{epsfig}
\usepackage{graphicx}
\usepackage{amsmath}
\usepackage{amssymb}

\usepackage[utf8]{inputenc}
\usepackage{color}
\usepackage{amsmath}
\usepackage{amsfonts}
\usepackage{fourier} 
\usepackage{array}
\usepackage{makecell}
\usepackage{changepage}
\usepackage{bigstrut}
\usepackage{multirow}
\usepackage{booktabs}  
\usepackage{graphicx}
\usepackage{subfigure}

\usepackage{authblk}
\usepackage{footnote}

\input{00def}

\usepackage[breaklinks=true,bookmarks=false]{hyperref}

\iccvfinalcopy 

\def\iccvPaperID{2240} 

\begin{document}

\title{Improving Adversarial Robustness via \GCE}


\newcommand*\samethanks[1][\value{footnote}]{\footnotemark[#1]}

\author[1]{Hao-Yun Chen\thanks{The authors contribute equally to this paper.}}
\author[1]{Jhao-Hong Liang\samethanks}
\author[1, 2]{Shih-Chieh Chang}
\author[3]{Jia-Yu Pan}
\author[3]{Yu-Ting Chen}
\author[3]{Wei Wei}
\author[3]{Da-Cheng Juan}

\affil[1]{\footnotesize Department of Computer Science, National Tsing-Hua University, Hsinchu, Taiwan}
\affil[2]{\footnotesize Electronic and Optoelectronic System Research Laboratories, ITRI, Hsinchu, Taiwan}
\affil[3]{\footnotesize Google Research, Mountain View, CA, USA}

\affil[ ]{\texttt{\{haoyunchen,jayveeliang\}@gapp.nthu.edu.tw}}
\affil[ ]{\texttt{{scchang}@cs.nthu.edu.tw}}
\affil[ ]{\texttt{\{jypan, yutingchen, wewei, dacheng\}@google.com}}

\maketitle

\input{00Abstract.tex}

\section{Introduction}
\label{sec:intro}
\input{01Introduction.tex}

\section{Related Work}
\label{sec:related work}
\input{02Related_Work.tex}

\section{\GCE}
\label{sec:GCE}
\input{03GCE.tex}

\section{Adversarial Setting}
\label{sec:02AdversarialSetting}
\input{02AdversarialSetting.tex}

\section{Experiments}
\label{sec:Experiments}

\input{04Experiments.tex}

\section{Conclusion}
\label{sec:Conclusion}
\input{05Conclusion.tex}

\clearpage

\clearpage



{\small
\bibliographystyle{ieee_fullname}
\bibliography{main}
}

\end{document}

%% file: 00def.tex
\newcommand{\GCE}{Guided Complement Entropy\xspace}
\newcommand{\BoldGCE}{{\bf G}uided {\bf C}omplement {\bf E}ntropy\xspace}
\newcommand{\abbrGCE}{GCE\xspace}

\newcommand{\CompEnt}{Complement Entropy\xspace}

\newcommand{\XE}{cross-entropy\xspace}
\newcommand{\abbrXE}{XE\xspace}

\long\def\hide#1{}

%% file: 00Abstract.tex
\begin{abstract}

Adversarial robustness has emerged as an important topic in deep learning as carefully crafted attack samples can significantly disturb the performance of a model. Many recent methods have proposed to improve adversarial robustness by utilizing adversarial training or model distillation, which adds additional procedures to model training. In this paper, we propose a new training paradigm called \textbf{G}uided \textbf{C}omplement \textbf{E}ntropy (GCE) that is capable of achieving ``adversarial defense for free,''  which involves no additional procedures in the process of improving adversarial robustness. In addition to maximizing model probabilities on the ground-truth class like \XE, we neutralize its probabilities on the incorrect classes along with a ``guided'' term to balance between these two terms. We show in the experiments that our method achieves better model robustness with even better performance compared to the commonly used \XE training objective. We also show that our method can be used orthogonal to adversarial training across well-known methods with noticeable robustness gain. To the best of our knowledge, our approach is the first one that improves model robustness without compromising performance.

\end{abstract}

%% file: 01Introduction.tex




Deep neural networks have been adopted to improve the performance of state-of-the-arts on a wide variety of tasks in computer vision, including image classification~\cite{krizhevsky2012imagenet}, segmentation~\cite{LinMBHPRDZ14}, and image generations~\cite{2014gan}. Albeit triumphing on predictive performance, recent literature~\cite{CarliniW16a, harnessing, PapernotMSW16} has shown that deep neural models are vulnerable to adversarial attacks. In an adversarial attack, undetectable but targeted perturbations are added to input samples which can drastically degrade the performance of a model. Such attacks have imposed serious threats to the safety and robustness of technologies enabled by deep neural models. Taking deep learning based self-driving cars, for example, models might mistakenly recognize a ``stop sign'' as a ``green light'' when adversarial examples are present. Needless to say, improving adversarial robustness is critical as it saves not only the model performance but the lives of people in many cases. 

\begin{figure}[!ht]
\begin{center}
\includegraphics[width=1.0\linewidth]{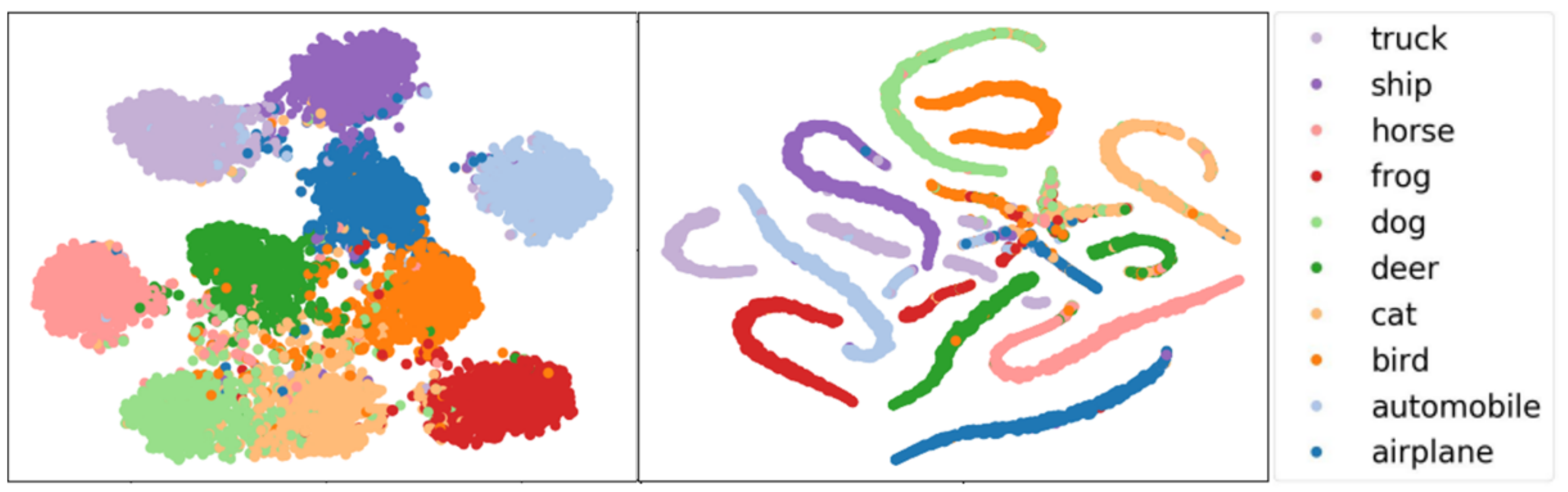}
\caption{Latent space of the models trained by different objective functions on CIFAR10. Visualization is done using t-SNE. Left: the latent space of model trained with cross-entropy (\abbrXE). Right: the latent space of model trained with \abbrGCE. Compared to \abbrXE, more distinct clusters (less overlap) are formed for each class from the training with \abbrGCE.}
\label{fig:tsne_normal_training_cifar10}
\end{center}
\end{figure}

A wide range of work has been proposed to address the issue of adversarial robustness. One method to improve the model robustness is ``adversarial training''~\cite{KurakinGB16a, madry2018towards, tramer2018ensemble} where the model is trained with either adversarial examples~\cite{madry2018towards} or a combination of both \textit{natural examples}\footnote{The ``natural examples'' mentioned in this paper are the normal samples in the original dataset, which contrasts to the ``adversarial examples.''} and adversarial examples as a form of data augmentation~\cite{tramer2018ensemble}. Here, adversarial examples refer to the artificially samples by adding targeted perturbations to the original data~\cite{CarliniW16a,Dong2018BoostingAA,harnessing,physical_word,madry2018towards,PapernotMSW16}. Other defense methods such as Defensive Distillation~\cite{PapernotM17, PapernotMWJS15} adopt the concept of model distillation to teach a smaller version of the original learned network that is less sensitive to input perturbations in order to make the model more robust. 

One caveat for using the existing defense mechanism is that they usually require additional processes, relying on either adversarial training or an additional teacher model in the distillation case. The fact that such a procedure is dependent on a specific implementation makes robustness improvement less flexible and more computationally intensive. The question we ask ourselves in this paper is, can we construct a training procedure that is capable of achieving ``adversarial defense for free,'' meaning that model robustness is improved in a model-agnostic way without the presence of an attack model or a teacher model. Another issue with the existing methods is that adversarial robustness usually come at the cost of model performance. A recent analysis~\cite{DBLP:journals/corr/abs-1901-00532, tsipras2018robustness} has shown that adversarial training hurts model generalization, and improvement on robustness is approximately at the same scale as the amount of performance degradation. 




In this paper, we propose a novel training paradigm to improve adversarial robustness that achieves adversarial defense for free without using additional training procedures. Specifically, we propose a carefully designed training objective called ``\BoldGCE'' (\abbrGCE). Different to the usual choice of \XE, which focuses on optimizing the model's likelihood on the correct class, we additionally add penalty that suppresses the model's probabilities on incorrect classes. Those two terms are balanced through a ``guided'' term that scales exponentially. Such a formulation helps to widen the gap in the manifold between ground-truth classes and incorrect classes, which has been proved to be effective in recent studies on \textit{minimum adversarial distortion}~\cite{weng2018evaluating}. This can be illustrated in Fig~\ref{fig:tsne_normal_training_cifar10} where \abbrGCE clearly makes the clusters more separable compares to \XE. Training with \abbrGCE for model robustness has several advantages compares to prior methods: (a) no additional computational cost is incurred as no adversarial example is involved and no extra model is required, and (b) contrary to prior analysis ~\cite{DBLP:journals/corr/abs-1901-00532, tsipras2018robustness} , improving model robustness no longer comes at a cost of model performance and we see sometimes better performance as supported in our experiment section.  



The contributions of our paper are three-fold. Firstly, to the best of our knowledge, \abbrGCE is the first work that achieves adversarial robustness without compromising model performance. Compares to the widely used methods which usually incur significant performance drop, our method can maintain or even beat the performance of  models trained with \XE. Secondly, our method is the first approach that is capable of achieving adversarial defense for free, which means improving robustness does not incur additional training procedures or computational cost, making the method agnostic to attack mechanisms. Finally, our proposed method managed to improve on top of a wide range of state-of-the-art defense mechanisms. Future work in the field can boost robustness improvements across different methods and push the frontiers of the adversarial defense forward.

%% file: 02Related_Work.tex
\paragraph{Adversarial Attacks.}


Several adversarial attack methods have been proposed in the ``white-box'' setting, which assumes the structure of the model being attacked is known in advance. As an iteration-based attack, ~\cite{harnessing} first introduces a fast method to crafting adversarial examples by perturbing the pixel's intensity according to the gradient of the loss function. ~\cite{harnessing} is an example of the \textit{single-step} adversarial attack. As an extension to ~\cite{harnessing}, ~\cite{physical_word} iteratively applies the gradient-based perturbation step by step, each with a small step size. A further extension to ~\cite{physical_word} is ~\cite{Dong2018BoostingAA} which adds gradient-based method with momentum to boost the success rates of the generated adversarial examples. In addition, an iterative method ~\cite{PapernotMSW16} has been proposed that uses Jacobian matrices to construct the saliency map for selecting pixels to modify at each iteration. As an optimization-based attack, the C\&W attack~\cite{CarliniW16a} is one of the most powerful attacks using the objective function to craft adversarial examples to fool the models.

\hide{  
}

\paragraph{Adversarial Defenses.}

Several defense strategies against adversarial attacks have been proposed to increase the model's robustness.  
In~\cite{KurakinGB16a}, the model's robustness is enhanced by using adversarial training on large scale models and datasets. \cite{madry2018towards} formulates the defense of model robustness as a min-max optimization problem, in which the adversary is constructed to achieve high loss value and the model is optimized to minimize the adversarial loss.  \cite{tramer2018ensemble} proposes an ensemble method which incorporates perturbed inputs transferred from other models, and yields model with strong robustness to black-box attacks. Besides improving the robustness by training with adversarial examples, Defensive Distillation~\cite{PapernotM17, PapernotMWJS15} is another effective defense approach. The idea is to generate a ``smooth" model which can reduce the sensitivity of the model to the perturbed inputs. In details, a ``teacher" model is proposed with a modified softmax function with a temperature constant. Then, using the \textit{soft} labels produced by the teacher network, a ``smooth'' model is trained and is found to be more resistant to adversarial examples.


\paragraph{Complement Objective Training.}


The proposed \GCE loss takes inspirations from 
{\bf C}omplement {\bf O}bjective {\bf T}raining (COT)~\cite{chen2018complement} which employs not only a primary loss function of \XE (\abbrXE), but also a ``complement'' loss function to achieve better generalization.  In COT, while the \abbrXE loss was to increase the output weight of the ground-truth class (and therefore, to learn to predict accurately), the ``complement'' loss function was designed with the \textit{intention} to neutralize the output weights on the incorrect classes (and therefore facilitates the training process and improves the final model accuracy). Although the complement loss function in COT was originally designed to make the ground-truth class stands out from the other classes, it has also been shown that the models trained using COT have achieved good robustness against \textit{single-step} adversarial attacks.

Despite the good robustness that COT achieved on single-step adversarial attacks, the two loss objectives that COT employs do not have a coordinating mechanism to efficiently work together to achieve robustness against stronger attacks, e.g., multiple-step adversarial attacks. 
We conjecture that 
the gradients from the two loss objectives may compete with each other and potentially compromise the improvements.

Based on the insight mentioned above, in this work, we propose \abbrGCE as an approach to reconcile the competition between the intentions of COT's two loss objectives. Rather than letting the two loss objectives work independently and coordinate merely 
via the normalization of output weights, our proposed \abbrGCE loss function unifies the core intentions of COT's two loss objectives, and explicitly formulates a mechanism to coordinate these core intentions. We argue that, by eliminating the competitions from COT's two loss objectives, the intention of "complement" loss can be maximumly expressed during the training phase to achieve better robustness.

%% file: 03GCE.tex


In this section, we introduce the proposed \GCE loss function, and discuss the intuition behind it.  We will first review the concept of \textit{Complement Entropy}~\cite{chen2018complement} before explaining the details of \abbrGCE.

\paragraph{Complement Entropy.}
\begin{table}[!h]
\begin{center}
\scalebox{0.7}{
\begin{tabular}[t]{ll}
Symbol  & Meaning                                                                                             \\ \hline
 $\hat{y_{i}}$       & \begin{tabular}[c]{@{}l@{}}The predicted probability for the $i$\textsuperscript{th} sample.\end{tabular} \\
$g$       & Index of the ground-truth class.                                                                    \\
 $y_{ij}$ or $\hat{y}_{ij}$ & The $j$\textsuperscript{th} class (element) of $y_i$ or $\hat{y}_i$.                                                                    \\
$N$ and $K$ & \begin{tabular}[c]{@{}l@{}}Total number of samples and total number of classes\end{tabular} 
\end{tabular}}
\end{center}

\caption{Basic Notations used in this section.}
\label{notations}
\end{table}
In~\cite{chen2018complement}, the Complement Entropy loss was introduced to facilitate the primary \XE loss during the training process.  It was shown that by introducing the Complement Entropy loss, the training process can generate models with better prediction accuracy as well as better robustness against single-step adversarial attacks.

\begin{align}
    -\frac{1}{N}\sum_{i=1}^{N}\sum_{j=1,j\neq g}^{K}( \frac{\hat{y}_{ij}}{1-\hat{y}_{ig}})\log( \frac{\hat{y}_{ij}}{1-\hat{y}_{ig}})
\label{eq:complement entropy}
\end{align}

Eq~(\ref{eq:complement entropy}) shows the mathematical formula of the Complement Entropy and notations are summarized in Table~\ref{notations}. We note that the idea behind the design of \CompEnt is to flatten out the weight distribution among the incorrect classes (``neutralize" the predicted weights on those classes).  Mathematically, a distribution is flattened when its entropy is maximized, so \CompEnt incorporates a negative sign to make it a loss function to be minimized.


Observing the results reported in~\cite{chen2018complement}, we argue that the improvement on the robustness of the model comes mostly from the property of the Complement Entropy on neutralizing the distributional weights on incorrect classes. Following this thought process, in this work, we formulate the property of the Complement Entropy explicitly into a new loss function that (a) is a standalone training objective with good empirical convergence behavior and (b) is explicitly designed to achieve robustness against various adversarial attacks (including both single-step and multi-step attacks).



\paragraph{\GCE.} Based on our observations mentioned above, we propose a novel training objective, \BoldGCE (\abbrGCE), which we will show that accomplishes our two original design goals: being a standalone training objective, and is inherently designed for achieving robustness against adversarial attacks.  Eq(\ref{eq:gce}) shows the mathematical formula of the proposed \abbrGCE:
\begin{align}
    -\frac{1}{N}\sum_{i=1}^{N} \hat{y}_{ig}^{\alpha}\sum_{j=1,j\neq g}^{K}( \frac{\hat{y}_{ij}}{1-\hat{y}_{ig}})\log( \frac{\hat{y}_{ij}}{1-\hat{y}_{ig}})
\label{eq:gce}
\end{align}


It can be seen that the Eq(\ref{eq:gce}) shares some similarity with the formula of the Complement Entropy in Eq(\ref{eq:complement entropy}), specifically the inner summation term, which we will call it the \textit{complement loss factor} of the \abbrGCE loss.  This similarity is intended, because it is our goal to make a loss function that explicitly takes advantage of the property of complement entropy on defending against adversarial attacks.  The main difference is that \abbrGCE also introduces a \textit{guiding factor} of $\hat{y}_{ig}^{\alpha}$ to modulate the effect of the complement loss factor, according to the model's prediction quality during the training iterations.  


\hide{ 
The intuition behind the formula of \abbrGCE is that, during the training iterations, the quality of the model may not start up to be very good, and the predicted value for the ground-truth class may include some arbitrariness during that time.  So, we argue that, during those early iterations, it is not strongly required to have the optimizer to optimize eagerly according to those loss values.  Intuitively, the proposed guiding factor is the control knob to modulate the amount of ``eagerness'' that the optimizer should treat the loss value.  

Mathematically, during the early training iterations, the prediction value on the ground-truth class ($\hat{y}_{ig}$) is not yet a high value, so the corresponding guiding factor $\hat{y}_{ig}^{\alpha}$ is also a smaller value, reducing the impact of the complement loss factor.  On the other hand, as the training process proceeds, the prediction quality of the model gradually improves and will assign larger values to the ground- truth class.  Consequently, the guiding factor will gradually increase the impact of the complement loss factor, which will encourage the optimizer to become more aggressive on neutralizing the weights on the incorrect classes, explicitly training towards a more robust model against adversarial attacks.
}

The intuition behind the formula of \abbrGCE is that, on a training instance where the predicted value for the ground-truth class is low, we consider that the model is not yet confident to its performance. So, we argue that, at this instance, it is not strongly required to have the optimizer to optimize eagerly according to the loss value.  Intuitively, the proposed guiding factor serves as the control knob that uses the predicted value for the ground-truth class to modulate the amount of ``eagerness'' that the optimizer should treat the loss value.  

Mathematically, on the instance that the model is not confident (when $\hat{y}_{ig}$ is small), the guiding factor $\hat{y}_{ig}^{\alpha}$ is also a small value, reducing the impact of the complement loss factor. On the other hand, as the model gradually improves and assigns larger values to the ground-truth class, the guiding factor will gradually increase the impact of the complement loss factor, which will encourage the optimizer to become more aggressive on neutralizing the weights on the incorrect classes, explicitly training towards a more robust model against adversarial attacks.





\paragraph{Analysis on the number of classes.}
The value of the proposed \abbrGCE loss as defined in Eq(\ref{eq:gce}) depends on the number of classes, $K$, of the learning task.  When using Eq(\ref{eq:gce}) directly in a training task, because the dynamic range of the training loss is different from that of other training tasks, additional efforts are needed on tuning the learning schedule for achieving good performance. 




Rather than using the \abbrGCE loss directly and fine-tuning the learning schedule for every training task, we mathematically divide the complement loss factor with a normalizing term $\log(K-1)$ to make the dynamic range of this \textit{normalized complement loss factor} between 0 and -1. We called the resulting loss function, the \textit{normalized \GCE} (Eq(\ref{eq:balance_gce})), which is defined as
\begin{align}
    -\frac{1}{N}\sum_{i=1}^{N} \hat{y}_{ig}^{\alpha}\cdot\frac{1}{\log(K-1)}\sum_{j=1,j\neq g}^{K}( \frac{\hat{y}_{ij}}{1-\hat{y}_{ig}})\log( \frac{\hat{y}_{ij}}{1-\hat{y}_{ig}})
\label{eq:balance_gce}
\end{align}
where $K$ is the number of classes for a training task.


By using the normalized \abbrGCE loss, we found that, without the extra effort of tuning the learning schedule, the optimizing algorithm can converge to a well-performing model, in terms of both the testing accuracy and the adversarial robustness. Based on this analysis, we conduct all of our experiments with the normalized \GCE in the following sections when we mention \abbrGCE.

\begin{figure*}[!h]
\begin{center}
\subfigure[Complement Entropy]
{\includegraphics[width=0.24\textwidth]{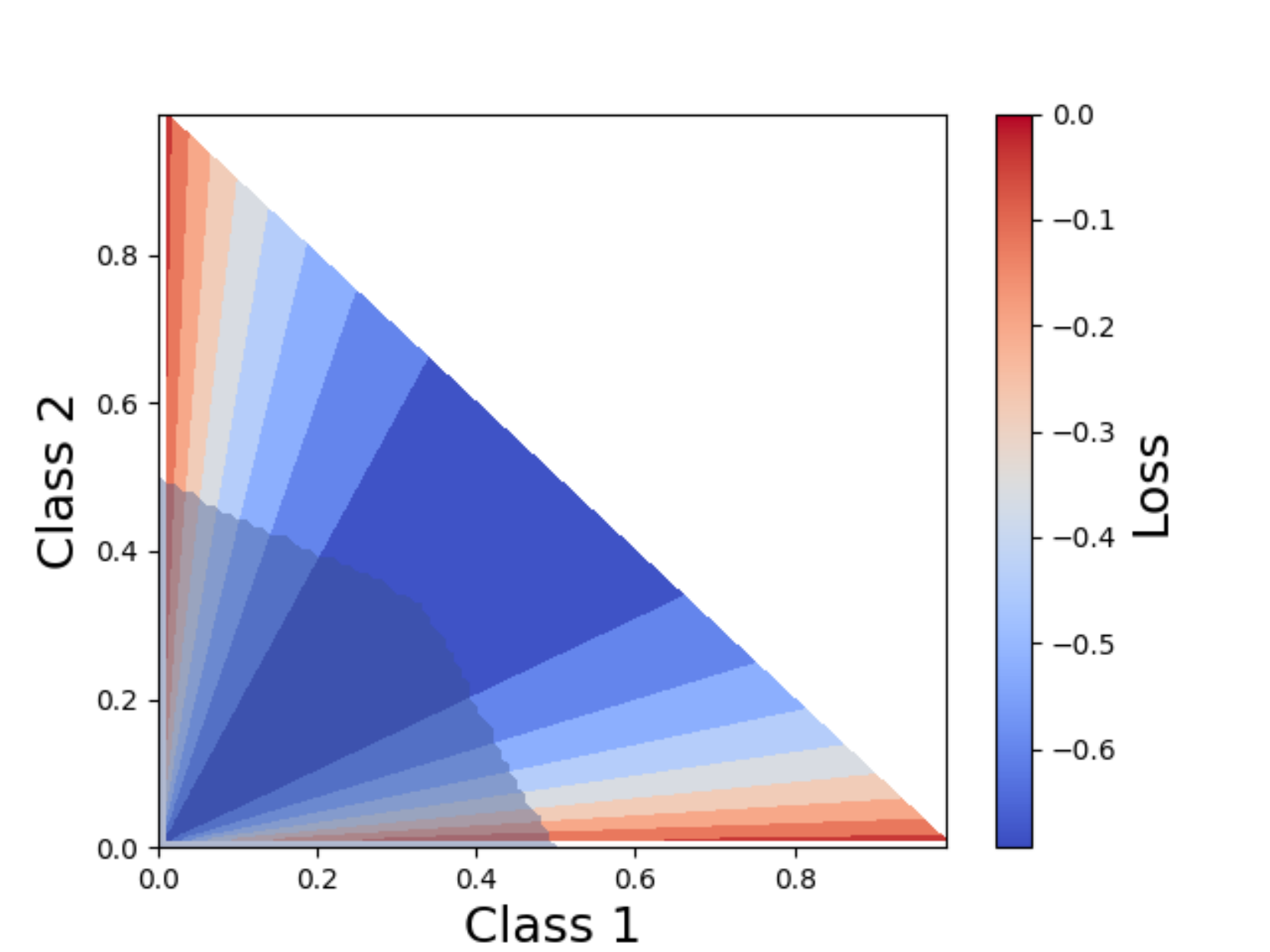}
\label{fig:Complement Entropy Synthentic}}
\subfigure[\abbrGCE with $\alpha = 1 $]
{\includegraphics[width=0.24\textwidth]{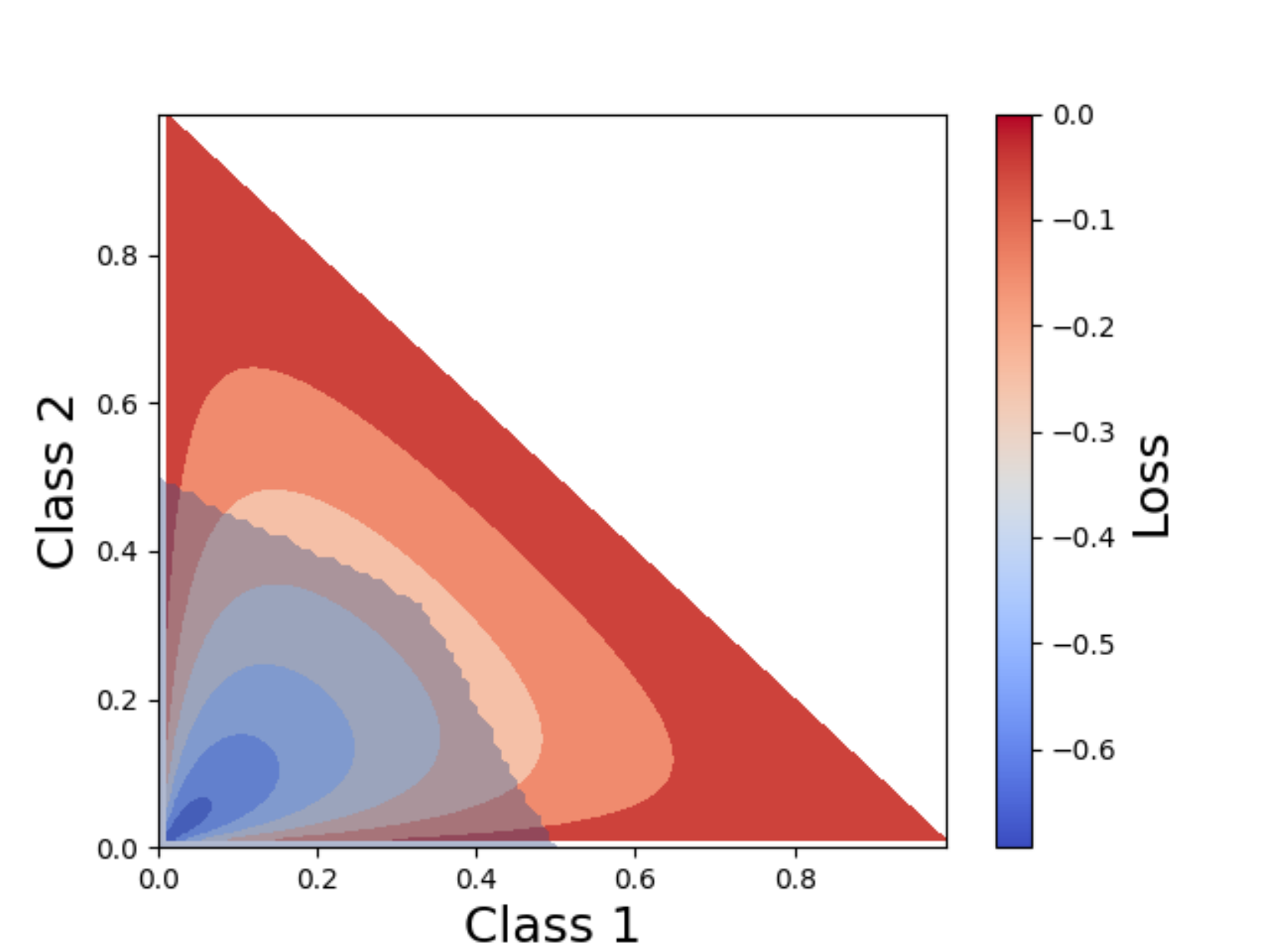}
\label{fig:GCE 1}}
\subfigure[\abbrGCE with $\alpha = 1/3 $]
{\includegraphics[width=0.24\textwidth]{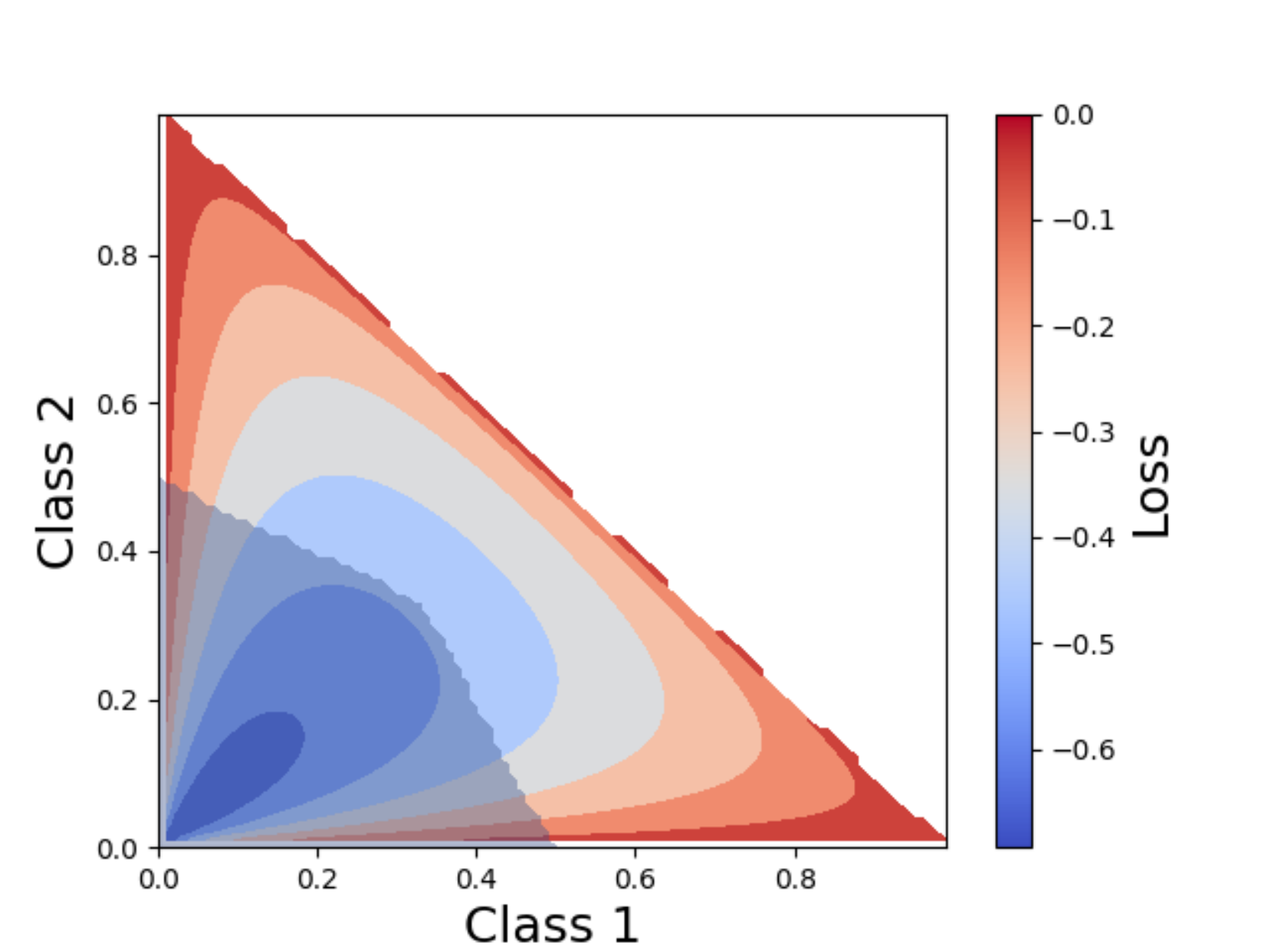}
\label{fig:GCE 3}}
\subfigure[\abbrGCE with $\alpha = 1/10 $]
{\includegraphics[width=0.24\textwidth]{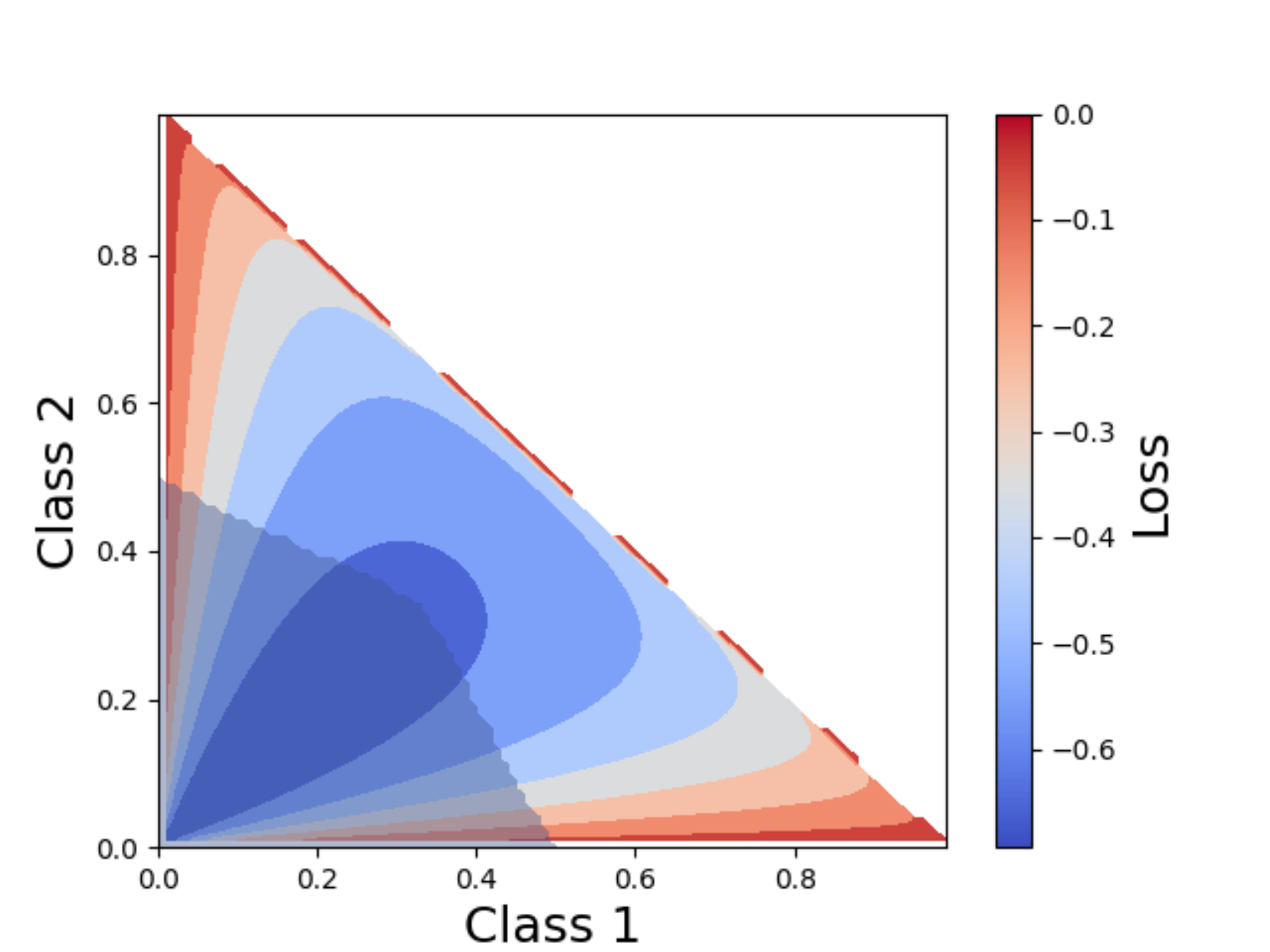}
\label{fig:GCE 10}}
\end{center}
\caption{Characteristics of \abbrGCE under different $\alpha$ values. Loss values are calculated assuming three classes, with class 0 being the ground-truth and class 1 \& 2 being incorrect classes. The X axis represents the predicted probability of class 1 and the Y axis for class 2. The shaded area (at the bottom left of each sub-figure) represents the prediction being correct (\ie, ground-truth class receives the predicted probability higher than class 1 or 2). Notice that in (a) and (d) the region of minimal loss (dark blue) does not overlap with the shaded region, which is not ideal as the loss function couldn't precisely reflect the prediction being correct. On the other hand, (b)(c) represent a preferred behavior of a loss function.}
\label{fig:synthetic_data_gce} 
\end{figure*}

\begin{figure}[!h]
\begin{center}
\includegraphics[width=0.6\linewidth]{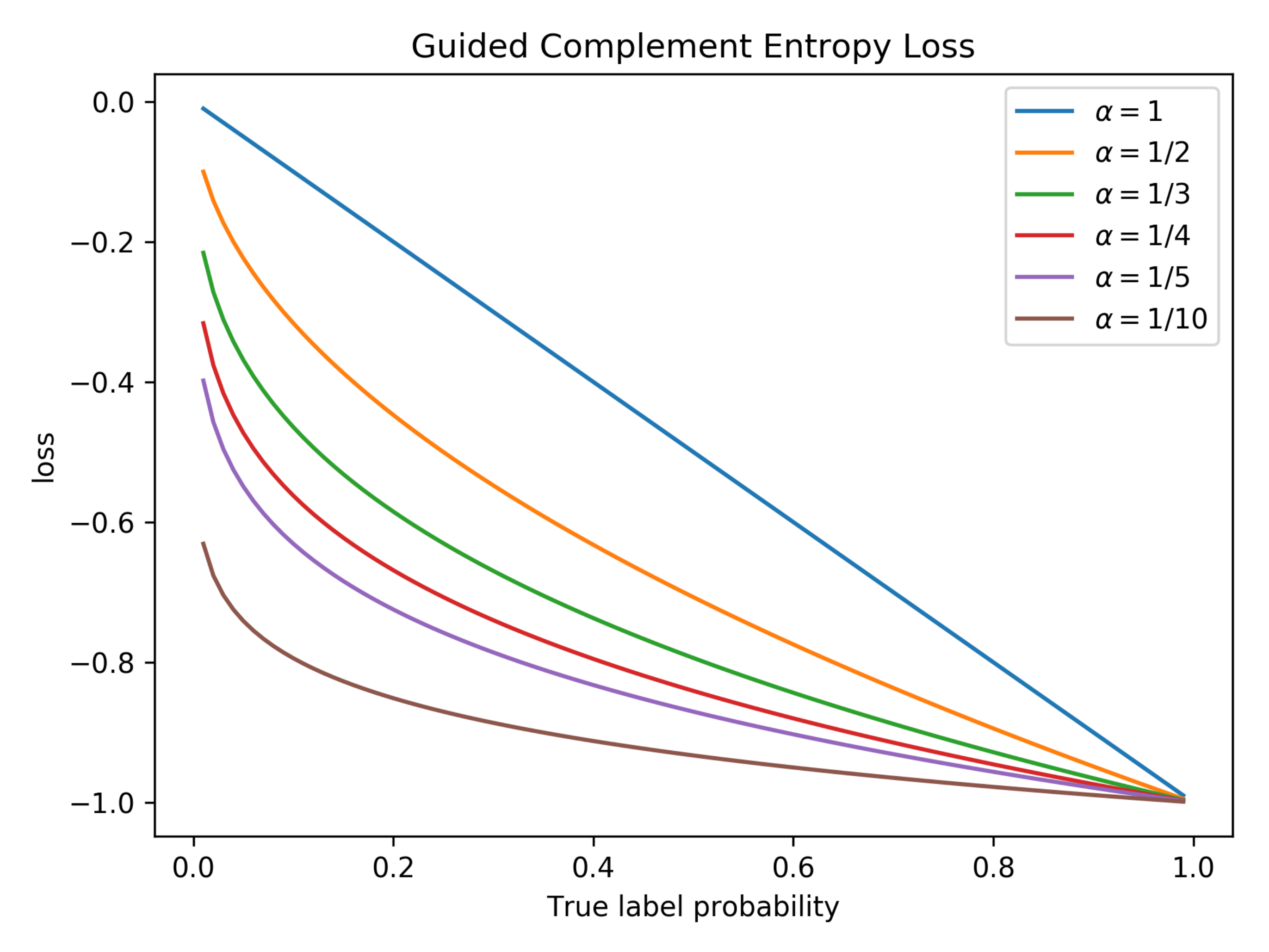}
\end{center}
\caption{The effect of the exponent $\alpha$ of the guiding factor, on our synthetic 3-class example data.  The X-axis is the output probability of the ground-truth class (class 0), so the optimal point is at the value $1.0$.  The output probability of the two incorrect classes are set to be equal (the optimal condition for the complement loss factor).  The Y-axis is the value of the \abbrGCE loss.  Different $\alpha$ values create curves of approaching slopes towards the optimal point.
}
\label{fig:gce loss with different exponential term}
\end{figure}

\paragraph{Synthetic Data Analysis.}

To further study the effect of the guiding factor of the \abbrGCE loss, $\hat{y}_{ig}^{\alpha}$, as well as how the exponent term $\alpha$ influences the loss function, we visualize the landscape of the \abbrGCE loss of a 3-class distribution and observe:
\begin{enumerate}
    \item How does the landscape of \abbrGCE loss differ to that of Complement Entropy loss?
    \item How does the $\alpha$ value modify the shape of the loss landscape of \abbrGCE?
    \item What are the implications on the convergence behavior, given the different loss landscapes of different $\alpha$ values?
\end{enumerate}

The synthetic training data we used in this exploratory study has only three classes, and we set that the class 0 be the ground-truth label, while the classes 1 and 2 are incorrect classes.   To visualize the landscape of a loss function over this 3-class synthetic data, we do a grid-sample over the weight distributions of the three classes, and plot the loss value on every sample point. 

Fig~\ref{fig:synthetic_data_gce} shows the visualization of the \CompEnt loss function over this synthetic distribution.  We note that for a three-class distribution, the loss function can be visualized on a \textit{2-D heat map}, where the X and Y coordinates indicate the values of two incorrect classes, and the heat-value corresponds to the value of the loss function. The value of the ground-truth class is uniquely determined by the two incorrect classes, since the three values have to sum up to $1$. Therefore, the origin point (0, 0) is the optimal point, since it is the point where the class 0 (the ground-truth class) gets the full probability value $1$.  Ideally, in this visualization, the loss values near the origin should be low, and the loss value increases as we move away from the origin.  In the figure, we also use a \textit{gray shade} to indicate the area around the origin where the output probability on the ground-truth class is larger than the output probabilities on the two incorrect classes. This gray-shade area is the region where the model will output the ground-truth class as its prediction.


In Fig~\ref{fig:Complement Entropy Synthentic}, although the loss value of \CompEnt is low around the origin, it can be seen that the loss values of all the points along the line $X=Y$ all have the equally low loss values.  When $X=Y$, the two incorrect classes have the same output probability values, and it can be shown in Eq~\ref{eq:complement entropy} that the \CompEnt loss value along line $X=Y$ is the same. Having a long "valley" in the loss landscape creates a problem when optimizing according to the \CompEnt loss.  The loss function only leads the training to converge to the valley, but not to the origin (the optimal point).

On the other hand, in Fig~\ref{fig:GCE 1}\subref{fig:GCE 3}\subref{fig:GCE 10}, it can be seen that the landscape of the \abbrGCE loss does not have the problematic "valley" in which the optimization process can be stuck.  By introducing the guiding factor, $\hat{y}_{ig}^{\alpha}$, in the \abbrGCE loss, the valley along the $X=Y$ line is no longer flat.  Instead, it is now a valley inclined downwards the origin (the model's optimal point).  Moreover, the downward slope of the valley is controlled by the exponent term, $\alpha$, of the guiding factor.  By comparing the different values of $\alpha$, it can be seen that a smaller $\alpha$ value, say $\alpha=1/10$, makes the loss value drops quickly to a low value, creating a larger ``basin" near the origin.

To further inspect the effect of the exponent $\alpha$, in Fig.~\ref{fig:gce loss with different exponential term}, we plot the profile of the \abbrGCE loss function along the line $X=Y$, when different $\alpha$ values is used. The X-axis is the output probability of the ground-truth class, so the value 1.0 is the optimal value.  When $\alpha=1$, the downward slope towards the optimal is a constant value. As the $\alpha$ value decreases, the downward slope has a bigger and bigger initial drop, followed by a shallower approach towards the optimal point. To maintain a good optimization momentum during the entire training process, our intuition is to prefer an $\alpha$ value that has a reasonable initial drop, but preserves a good final approach as we get closer to the optimal point.  Therefore, we argue that the preferable $\alpha$ value should not be too close to $1$, nor be too small.  In our experiments in the following section, we will try multiple values of $\alpha$ and report the results.

%% file: 02AdversarialSetting.tex

In the adversarial setting, adversaries apply attacking methods to craft adversarial examples based on the given natural examples. We consider the white-box attack, which is the most challenging and difficult threat model for classifiers to
defend~\cite{CarliniW16a}. White-box adversarial attacks
know everything, e.g., parameters, about the models that they attack on. The $\epsilon$ mentioned below is the perturbation for the adversarial attacks.

\paragraph{Fast Gradient Sign Method (FGSM)~\cite{harnessing}} introduced an efficient one-step attack. The method uses the gradient evaluated by the training cost function to
determine the direction of the perturbation.The adversarial examples $x^*$ can be simply generated by :
\begin{equation}
\mathit{x^*}= \mathit{x} + \epsilon \cdot sign( \,\triangledown_{\mathit{x}} \mathit{L}( \,\mathit{x},\mathit{y} )\,)\,
\end{equation}
,where the $\epsilon$ is the perturbation and $\mathit{L}(\mathit{x},\mathit{y} )$ is the training loss function.

\paragraph{Basic Iterative Method (BIM)~\cite{physical_word}} introduced the extension of FGSM which applies multiple steps perturbation and clipped the value of features in the constrained bounding. The BIM formulation is :
\begin{equation}
\mathit{x_{0}^{*}} = \mathit{x}, \quad        \mathit{x_{i}^{*}} = clip_{\mathit{x}, \epsilon}( \,\mathit{x_{i-1}^{*}} + \frac{\epsilon}{\mathit{r}} \cdot sign( \,\triangledown_{\mathit{x_{i-1}^{*}}} \mathit{L}( \,\mathit{x_{i-1}^{*}},\mathit{y} )\,)\,)\,
\end{equation}
where the $\mathit{r}$ is the number of iterations and $\mathit{clip_{\mathit{x,\epsilon}}}(\cdot)$ is the clipping function to keep the value of features being bounded.

\paragraph{Projected Gradient Descent (PGD)~\cite{madry2018towards}} proposed a more powerful adversary method which is the multi-step variant $\mathit{FGSM^{k}}$. The process of crafting the adversarial examples in PGD is similar to BIM. The difference is that the $\mathit{x_{0}^{*}}$ is a uniformly random point in $\ell_{\infty}$-ball around $\mathit{x}$.

\paragraph{Momentum Iterative Method (MIM)~\cite{Dong2018BoostingAA}}
 integrated the momentum property into the iterative gradient-based attack to craft the adversarial examples. The method not only stabilize the update directions during the iterative process but also improve the situation about sticking in the local maximum in BIM. The MIM formulation is:
 \begin{equation}
 \mathit{g_{t}} = \mu \cdot \mathit{g_{t-1}} + \frac{\triangledown_{\mathit{x}} \mathit{L}( \,\mathit{x_{t-1}^{*}},\mathit{y} )\,}{\|\triangledown_{\mathit{x}} \mathit{L}( \,\mathit{x_{t-1}^{*}},\mathit{y} )\,\|_{1}}
 \end{equation}
 \begin{equation}
 \mathit{x_{t}^{*}} = clip_{\mathit{x}, \epsilon}( \,\mathit{x_{t-1}^{*}} + \frac{\epsilon}{\mathit{r}} \cdot sign(\mathit{g_{t}})\,)\,
 \end{equation}
 where $\mathit{g_{t}}$ is the gradient which  accumulating the velocity vector in the direction and $\mu$ is the decay factor.
\paragraph{Jacobian-based Saliency Map Attack (JSMA)~\cite{PapernotMSW16}}
proposed the powerful target attack which can just perturb fewer pixels. The method identify the features that can significantly affect output classification by the evaluation of the saliency map. Through modifying the input features iterative, JSMA craft the adversarial example which cause the model misclassified in specific targets.


\paragraph{Carlini \& Wagner (C\&W) ~\cite{CarliniW16a}} introduce a optimization-based attack and can effective defeat defensive distillation~\cite{CarliniW16a}. To ensure the perturbation for images is available, the method defines the box constraints to make the pixels value in a constrained bounding. They define:
\begin{equation}
\mathit{x^{*}} = \frac{1}{2}(\,\mathit{tanh(\,w)\,} + 1 )\,
\end{equation}
in terms of $\mathit{w}$ and let $0 \leq \mathit{x}^{*} \leq 1$ to make the sample is valid and optimize $\mathit{w}$ with the formulation:
\begin{equation}
\min_{\mathit{w}}\|\frac{1}{2}(\,\mathit{tanh(\,w)\,} + 1 )\,- \mathit{x}\|_{2}^{2} + \mathit{c} \cdot \mathit{f}(\,\frac{1}{2}(\,\mathit{tanh(\,w)\,} + 1 )\,)\,
\end{equation}
where  $\mathit{c}$ is the constant. The $\mathit{f(\,\cdot)\,}$ is the objective function 
\begin{equation}
\mathit{f(\,x)\,} = \max(\max\{\mathit{Z_{pre}(x)_{i}} : i \neq \mathit{y}\} - \mathit{Z_{pre}(x)_{i}}, -\kappa)
\end{equation} 
where the $\kappa$ is the confidence and $\mathit{Z_{pre}(x)_{i}}$ is the model output logits.

%% file: 04Experiments.tex
We conduct experiments to demonstrate that:
\begin{enumerate}
    \item Models trained with \abbrGCE can achieve better classification performance, compared to the baseline models trained using the \abbrXE loss function.
    \item In addition to achieving good classification performance on the natural, non-adversarial examples, the models trained with \abbrGCE are also robust against several kinds of "white-box" adversarial attacks.
    \item In the setting of "adversarial training", we show that substituting the \abbrGCE loss function in the PGD adversarial training, the resulting models are more robust than the previous results.
\end{enumerate}

\subsection{Performance on natural examples}
\label{sec:accuracy}

In this section, we give experimental results showing that models trained using \abbrGCE, in the natural, non-adversarial setting, can outperform the previously reported best models trained using \abbrXE.  
Specifically, we compared the model accuracy on several image classification datasets of different scales, ranging from MNIST~\cite{726791}, CIFAR10, CIFAR100~\cite{Krizhevsky09learningmultiple} and Tiny ImageNet\footnote{\url{https://tiny-imagenet.herokuapp.com}, a subset of ImageNet~\cite{imagenet_cvpr09}}. 

In our experiments, for each data set, we take the best model previously published (the baseline model), and substitute the loss function from the original \abbrXE to the proposed \abbrGCE.  For MNIST, we use the model Lenet-5~\cite{726791} with Adam Optimizer. For CIFAR10 and CIFAR100, we use ResNet-56~\cite{He_2016_CVPR}; while for Tiny ImageNet, it is trained with ResNet-50.  The ResNet-56 and ResNet-50 models were trained following the standard settings described in~\cite{He_2016_CVPR}.  In details, the models were trained using SGD optimizer with momentum of 0.9, and weight decay is set to be 0.0001.  The learning rate is set to start at 0.1, then is divided by 10 at the 100\textsuperscript{th} and 150\textsuperscript{th} epochs.  

Table~\ref{tb:Convergence Classification error rate} compares the classification error rates of the baseline models and those of \abbrGCE's models.  We found that the performance achieved by \abbrGCE's models are usually as good or outperforming the models from \abbrXE, when the guided factor, controlled by $\alpha$, is appropriately chosen.  For example, on Tiny ImageNet, our proposed model achieves 38.56\% error rate, at $\alpha = 1/3$, which is better than the 39.54\% error rate of the baseline model.


\begin{table}[!h]
\begin{center}
\scalebox{0.8}{
\begin{tabular}{|c|c|c|c|c|}
\hline

Dataset      & MNIST   & CIFAR10 & CIFAR100          & Tiny ImageNet \\ \hline
Architecture & LeNet-5 &\multicolumn{2}{c|}{ResNet-56} & ResNet-50       \\ \hline \hline
Baseline            & 0.8           & 7.99        & 31.9         & 39.54   \\
$\alpha = 1/2$      & \bf0.61       & 9.18        & 40.59         & 43.36       \\
$\alpha = 1/3$      & \bf0.67       & \bf7.18     & \bf31.75         &\bf38.56\\
$\alpha = 1/4$      & \bf0.64       & \bf6.93     & \bf31.8         & \bf38.69      \\
$\alpha = 1/5$      & \bf0.68       & \bf6.91     & \bf31.48         & \bf38.26      \\ \hline
\end{tabular}
}
\end{center}
\caption{Classification error rates (\%) of the baseline models (using \abbrXE) and the proposed models, evaluated on 4 image classification data sets. The $\alpha$ is the guided factor of the proposed model.}
\label{tb:Convergence Classification error rate}
\end{table}

\subsection{Robustness to White-box attacks}

The main motivation of the proposed \abbrGCE loss, is to train models that are robust to adversarial attacks.  In this section, we took the models that were trained above as described in Sec.~\ref{sec:accuracy}, and evaluated their robustness against the six kinds of white-box attack (described in Sec.~\ref{sec:02AdversarialSetting}). In the experiments of this section, we set the exponent of the guiding factor $\alpha = 1/3$.


\paragraph{Robustness.}
We first evaluated the robustness of our proposed models on the two smaller data sets, MNIST and CIFAR10.
Following the preprocessing that is common in previous work, the pixel values in both data sets were scaled to the interval $[0, 1]$.  For the iteration-based attacks using gradients, e.g., FGSM, BIM, PGD and MIM, we crafted adversarial examples in the non-targeted way, with respect to the perturbation $\epsilon$. The iterations are set to be 10 for BIM, and 40 for both PGD and MIM.  For the iteration-based attack using Jacobian matrix, JSMA, the adversarial examples were perturbed with the several values of $\gamma$ (the maximum percentage of pixels perturbed in each image), and the perturbation $\epsilon = $1.  For the optimization-based attack, C\&W, we perform the targeted attack using the "average case" approach, as mentioned in the original paper~\cite{CarliniW16a}.  Regarding the parameters of the C\&W attack, we set binary steps to be 9 and the maximum iterations to be 1000. The initial constant is set to 0.001 and the confidence is set to 0. 


Table~\ref{main table for adversairal attack on noraml training} shows the results of the attacks mentioned above.  The models trained with \abbrGCE always have higher classification accuracy than the baseline models trained with \abbrXE, under the six white-box adversarial attacks, on both datasets. In particular, the best accuracy improvement between our models and the baseline is on the Momentum Iterative Method (MIM) attack.


For large-scale datasets, i.e., CIFAR100 and Tiny ImageNet, we evaluated the robustness of our models on the PGD attack, which is the most powerful white-box adversarial attack.  Table~\ref{tb:Performance on white-box adversarial attacks in CIFAR-100 and Tiny Imagenet} compares the classification accuracy of our models and that of the baseline models. Our models, under the PGD attack, outperform the baseline models on classification accuracy.


\begin{table}[!h]
\begin{center}
\scalebox{0.8}
{
\begin{tabular}{|c|c|c|c|c|c|c|}
\hline
\multirow{2}{*}{Attacks} & \multicolumn{3}{c|}{\bf MNIST}                                                                                                                                            & \multicolumn{3}{c|}{\bf CIFAR10}                                                                                                                                         \\
                         & Param.                                 
                         & \abbrXE                     
                         & \bf \abbrGCE                           
                         & Param.                            
                         & \abbrXE                           
                         & \bf \abbrGCE                                               \\ \hline \hline
FGSM                     & 
\begin{tabular}[c]{@{}c@{}}$\epsilon= 0.1$ \\ $\epsilon = 0.2$\\ $\epsilon = 0.3$\end{tabular}  & 
\begin{tabular}[c]{@{}c@{}}78.32 \\ 38.88 \\ 14.99\end{tabular} & \begin{tabular}[c]{@{}c@{}}\bf87.66\\ \bf62.74\\ \bf47.21\end{tabular} & \begin{tabular}[c]{@{}c@{}}$\epsilon= 0.04$ \\ $\epsilon = 0.12$\\ $\epsilon = 0.2$\end{tabular} & 
\begin{tabular}[c]{@{}c@{}}14.76\\ 9.58\\ 8.78\end{tabular} & \begin{tabular}[c]{@{}c@{}}\bf41.22\\ \bf14.82\\ \bf11.81\end{tabular} \\ \hline

BIM                      & 
\begin{tabular}[c]{@{}c@{}}$\epsilon= 0.1$ \\ $\epsilon = 0.2$\\ $\epsilon = 0.3$\end{tabular} & 
\begin{tabular}[c]{@{}c@{}}53.14\\ 2.15\\ 0.01\end{tabular} & \begin{tabular}[c]{@{}c@{}}\bf61.92\\ \bf34.49\\ \bf33.45\end{tabular} & \begin{tabular}[c]{@{}c@{}}$\epsilon= 0.04$ \\ $\epsilon = 0.12$\\ $\epsilon = 0.2$\end{tabular}      & 
\begin{tabular}[c]{@{}c@{}}0.25\\ 0.0\\ 0.0\end{tabular} & \begin{tabular}[c]{@{}c@{}}\bf19.59\\ \bf3.03\\ \bf1.97\end{tabular} \\ \hline
PGD                      & 
\begin{tabular}[c]{@{}c@{}}$\epsilon= 0.1$ \\ $\epsilon = 0.2$\\ $\epsilon = 0.3$\end{tabular}       & 
\begin{tabular}[c]{@{}c@{}}46.85\\ 1.58\\ 0.0\end{tabular} & \begin{tabular}[c]{@{}c@{}}\bf51.85\\ \bf9.55\\ \bf2.22\end{tabular} & \begin{tabular}[c]{@{}c@{}}$\epsilon= 0.04$ \\ $\epsilon = 0.12$\\ $\epsilon = 0.2$\end{tabular}       & 
\begin{tabular}[c]{@{}c@{}}0.0\\ 0.0\\ 0.0\end{tabular} & \begin{tabular}[c]{@{}c@{}}\bf5.91\\ \bf1.89\\ \bf1.66\end{tabular} \\ \hline

MIM                      & 
\begin{tabular}[c]{@{}c@{}}$\epsilon= 0.1$ \\ $\epsilon = 0.2$\\ $\epsilon = 0.3$\end{tabular}       & 
\begin{tabular}[c]{@{}c@{}}48.28\\ 2.29\\ 0.01\end{tabular} & \begin{tabular}[c]{@{}c@{}}\bf61.18\\ \bf39.81\\ \bf38.78\end{tabular} & \begin{tabular}[c]{@{}c@{}}$\epsilon= 0.04$ \\ $\epsilon = 0.12$\\ $\epsilon = 0.2$\end{tabular}       & 
\begin{tabular}[c]{@{}c@{}}0.0\\ 0.0\\ 0.0\end{tabular} & \begin{tabular}[c]{@{}c@{}}\bf15.44\\ \bf13.1\\ \bf12.69\end{tabular} \\ \hline
JSMA                     & 
\begin{tabular}[c]{@{}c@{}}$\gamma = 0.25$\\$\gamma = 0.5$\end{tabular}             & \begin{tabular}[c]{@{}c@{}}1.53\\ 0.1\end{tabular}     & \begin{tabular}[c]{@{}c@{}}\bf26.24\\ \bf17.26\end{tabular}     & \begin{tabular}[c]{@{}c@{}}$\gamma = 0.07$\\ $\gamma = 0.14$\end{tabular}             & \begin{tabular}[c]{@{}c@{}}1.09\\ 0.14\end{tabular}     & \begin{tabular}[c]{@{}c@{}}\bf18.72\\ \bf10.94 \end{tabular}     \\ \hline
C\&W    & $c = 0.$  & 0.0  & \bf25.6    & $c = 0.$  & 0.0 & \bf0.8                                                 \\ \hline
\end{tabular}
}
\end{center}
\caption{Performance (\%) on white-box adversarial attacks with wide range of perturbations. The model for MNIST is Lenet-5 and CIFAR10 is Resnet-56. For FGSM, BIM, PGD and MIM, we select three perturbations, $\epsilon = $ 0.04, 0.12 and 0.2 in our experiment. In JSMA, we set the perturbation $\epsilon = 1.$ and maximum iterations to be 100 and 200, which means the maximum pixels that JSMA purturbs in each image. We show the max iteration which is transformed to percentage of maximum pixels modified $\gamma$ in our experiment. In C\&W, we set confidence $c = 0.$ and maximum iterations is 1000.}
\label{main table for adversairal attack on noraml training}
\end{table}


\begin{table}[!h]
\begin{center}
\scalebox{0.8}{
\begin{tabular}{|c|c|c|c|c|c|c|}
\hline
\multirow{2}{*}{Attacks} & \multicolumn{3}{c|}{CIFAR100}                                                                                                                    & \multicolumn{3}{c|}{Tiny ImageNet}                                                                                                                                   \\
                         & Parameter                                    
                         & \abbrXE                                           
                         & \abbrGCE                                          
                         & Parameter                                    
                         & \abbrXE                                       
                         & \abbrGCE                                               \\ \hline
PGD                      & 
\begin{tabular}[c]{@{}c@{}}$\epsilon = 0.04$\\$\epsilon = 0.12$\\ $\epsilon = 0.2 $\end{tabular} & 
\begin{tabular}[c]{@{}c@{}}0.04\\ 0.0\\ 0.0\end{tabular} & \begin{tabular}[c]{@{}c@{}}\bf 2.94\\ \bf0.46\\ \bf0.19\end{tabular} & \begin{tabular}[c]{@{}c@{}}$\epsilon = 0.04$\\$\epsilon = 0.12$\\ $\epsilon = 0.2 $\end{tabular} & 
\begin{tabular}[c]{@{}c@{}}0.0\\ 0.0\\ 0.0\end{tabular} & \begin{tabular}[c]{@{}c@{}}\bf9.52\\ \bf4.27\\ \bf1.11\end{tabular} \\ \hline

\end{tabular}
}
\end{center}
\caption{ Performance (\%) on white-box adversarial attacks with wide range of perturbations in CIFAR100. The model of CIFAR100 and Tiny ImageNet is Resnet-56.}
\label{tb:Performance on white-box adversarial attacks in CIFAR-100 and Tiny Imagenet}
\end{table}




\paragraph{Robustness compared to COT.}
\label{White-box adversarial attacks compared to COT}

For evaluating the adversarial robustness about COT and \abbrGCE, we conduct various white-box attacks on the models trained with COT and \abbrGCE with different perturbations in MNIST and CIFAR10. In Table~\ref{robustness_COT}, we show that the accuracy of various adversarial attacks trained with \abbrGCE outperforms COT.

\begin{table}[!h]
\begin{center}
\scalebox{0.7}{
\begin{tabular}{|c|c|c|c|c|c|c|c|c|}
\hline
\multirow{2}{*}{Attacks} & \multicolumn{4}{c|}{MNIST} & \multicolumn{4}{c|}{CIFAR 10} \\
& Param.  & XE  & COT  & GCE & Param.   & XE   & COT  & GCE  \\ \hline
FGSM                     
& $\epsilon$ = 0.2     &38.88     &   51.8   & \bf62.74    & $\epsilon$ = 0.04     &14.76      & 33.62     & \bf41.22     \\ \hline
BIM                      
& $\epsilon$ = 0.2     &2.15     &  4.35    &  \bf34.49   & $\epsilon$ = 0.04     &0.25      & 7.49     &  \bf19.59    \\ \hline
MIM                      
& $\epsilon$ = 0.2     &2.29     &  4.26    & \bf39.81    & $\epsilon$ = 0.04     &0.0      & 0.0     &  \bf15.44    \\ \hline
JSMA                     
& $\gamma$ = 0.25    &1.53     &  11.13    &  \bf26.24   & $\gamma$ = 0.07     &1.09      & 8.25      &   \bf18.72   \\ \hline
C\&W                     
& c = 0       &0.0     &   11.9   & \bf25.6    & c = 0        &0.0      &   0.0   &   \bf0.8   \\ \hline
\end{tabular}}
\end{center}
\caption{Performance (\%) under various white-box adversarial attacks between COT and GCE across MNIST and CIFAR10.}
\label{robustness_COT}
\end{table}

\subsection{Robustness to adversarial training}

The idea of adversarial training is to include adversarial examples in the training phase, to create models that are robust to other adversarial examples during the test phase. Several frameworks of adversarial training have been proposed.  In this work, we choose to integrate our proposed \abbrGCE loss function in the Projected Gradient Descent (PGD) adversarial training, since the PGD attack is considered an universal one, among all of the first-order adversarial attacks~\cite{madry2018towards}.  We show that the resulting models from this integration are more robust than the ones trained using the original PGD approach.



The PGD adversarial training uses a min-max objective function to accomplish adversarial training: 
\begin{equation}
    \underset{\theta}{\min}~\rho(\theta), \;\text{where} \;\rho(\theta) =  \mathop{\mathbb{E}}_{x,y\sim D}[ \underset{\delta}{\max}~\mathit{L}(\theta, x + \delta, y)].
\label{eq:pgd_min_max}
\end{equation}
, where $\mathit{D}$ is the data distribution over pairs of training sample $\mathit{x}$ and label $\mathit{y}$.  The loss function $\mathit{L(\cdot)}$ is the \abbrXE loss.  In Eq(\ref{eq:pgd_min_max}), the inner maximization problem is for crafting training adversarial examples to induce maximum loss values, while the outer minimization problem is for building a classification model, $\rho(\cdot)$, to minimize the adversarial loss by the universal adversary.  One typical approach for optimizing this min-max objective is through an iterative algorithm.



In the original work, the loss function for the inner maximization and that for the outer minimization are the same, which is the \abbrXE loss.  In our work, we keep the loss function for the inner maximization intact as the \abbrXE loss, because it has been proved that the PGD framework generates the optimal adversarial examples, among all first-order adversarial attacks, when using the \abbrXE loss.  On the other hand, for the outer minimization, that is, the training of the classification model, we replace the \abbrXE loss with our proposed \abbrGCE loss.  This way of integration is similar to other previous work \cite{raghunathan2018certified} that also keeps the \abbrXE as the loss function of the inner maximization problem.


In our setup, we use \abbrGCE ($\alpha = 1/3$) as the loss function for the Empirical Risk Minimization (ERM)~\cite{vapnik1999overview} instead of original \abbrXE in Eq(\ref{eq:pgd_min_max}). Then, to compare the robustness of the models generated using our setup, with the models trained using the original setup, we attack both these models using the PGD white-box (with respect to the \abbrXE loss) adversarial attack.



In our experiments, the minimization models we used are the baseline models as described in the previous sections, i.e., LeNet or Resnet, for their corresponding datasets. 
Table~\ref{Robustness of Adversarial training under PGD adversarial attacks on MNIST and CIFAR-10} shows the comparison results on MNIST and CIFAR10 datasets.
More specifically, in our experiments, we use the same settings of the iterative optimization as used in the previous work~\cite{madry2018towards}, to conduct the adversarial training and adversarial attacks: on MNIST, we do 40 iterations of crafting adversarial examples during training; at the testing phase, 100 iterations are used to apply the PGD attack.  On CIFAR10, 10 iterations of adversarial training are used, and the adversarial attacks are conducted with 40 iterations. We demonstrate better robustness while using \abbrGCE loss for the outer minimization.


\begin{table}[!h]
\begin{center}
\scalebox{0.7}{
\begin{tabular}{|c|c|c|c|c|c|c|}
\hline
\multirow{2}{*}{Attacks} & \multicolumn{3}{c|}{MNIST} & \multicolumn{3}{c|}{CIFAR10}                                                                                                                 \\
                         & perturbation     & \abbrXE     & \abbrGCE    & perturbation                                          & \abbrXE                                            & \abbrGCE                                           \\ \hline \hline
PGD                      
& $\epsilon = 0.3$ & 83.67 & \bf83.85    & 
\begin{tabular}[c]{@{}c@{}}
$\epsilon = 0.04$\\ $\epsilon = 0.08$\end{tabular} & \begin{tabular}[c]{@{}c@{}}
41.50\\ 12.93\end{tabular} & \begin{tabular}[c]{@{}c@{}}
\bf41.57\\ \bf13.16\end{tabular} \\ \hline
\end{tabular}
}
\end{center}
\caption{Performance (\%) of Adversarial training under PGD adversarial attacks on MNIST and CIFAR10.}
\label{Robustness of Adversarial training under PGD adversarial attacks on MNIST and CIFAR-10}
\end{table}



\begin{figure}[!h]
\begin{center}
\includegraphics[width=1.0\linewidth]{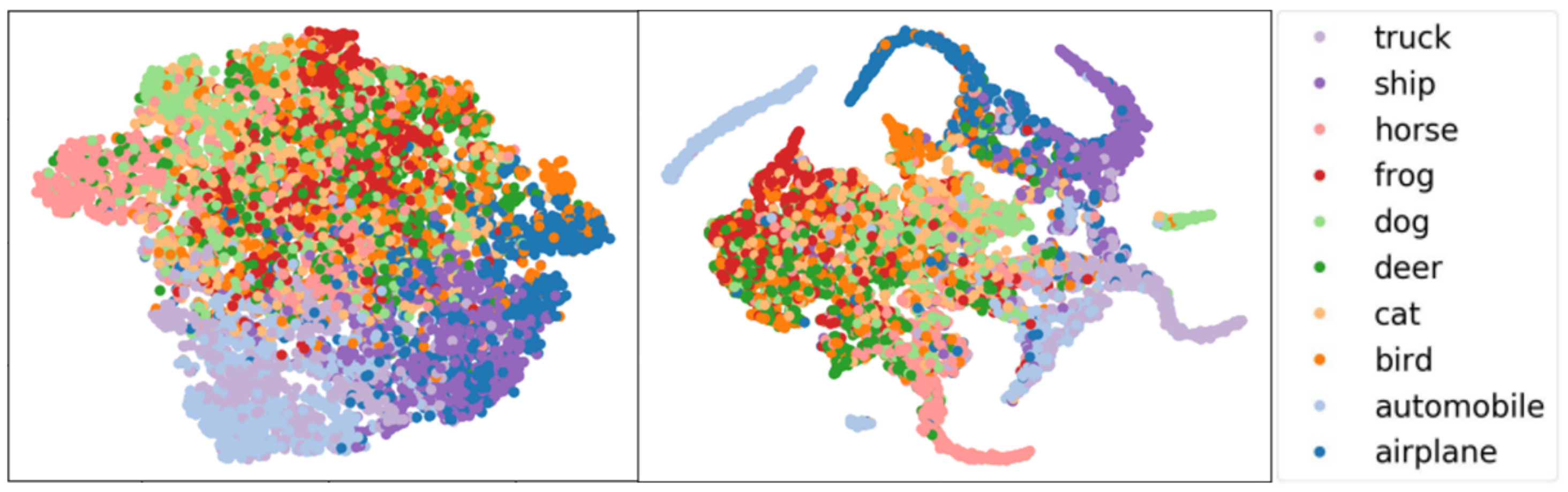}
\caption{Latent spaces of the adversarial-trained models on CIFAR10: (Left) latent space of \abbrXE's model; (Right) latent space of \abbrGCE's model. The adversarial training is done using PGD with $\epsilon=0.02$. Visualization is done using t-SNE.}
\label{fig:tsne_adversarial_training_cifar10}
\end{center}
\end{figure}

\paragraph{Latent space of adversarial trained models.}
We also inspect the latent spaces of \abbrGCE's models trained with PGD adversarial training, and find that they have similar characteristics of the latent spaces of the \abbrGCE's models from the natural training\footnote{We use "natural training" to refer to the training process using only the natural examples in the original dataset, in contrast to adversarial training that takes adversarial examples during training.} procedure. For example, in Fig~\ref{fig:tsne_adversarial_training_cifar10}, we visualize the latent space of our model trained on the CIFAR10 dataset.  It can be seen that, despite the presence of many adversarial training examples, our model is still able to disperse examples of different classes and create visually better separated clusters.




%% file: 05Conclusion.tex
In this paper, we present the {\bf G}udied {\bf C}omplement {\bf E}ntropy (GCE), a novel training objective to answer the motivational question: \textit{``how to improve the model robustness, and at the same time, keep or improve the performance when no attack is present?''} \abbrGCE encourages models to learn latent representation that groups samples of the same class into distinct clusters. Experimental results show that, under the normal condition (no adversarial attack is present), the state-of-the-art models trained with \abbrGCE achieves better accuracy compared to cross-entropy by up to relative 10.14\% on CIFAR-10. When adversarial attacks are present, experimental results show that models trained with \abbrGCE are more robust compared to \abbrXE. Under PGD attacks, \abbrGCE outperforms the baseline with improvement up to 5.91\%. Our experimental results also confirm that \abbrGCE can be combined with PGD adversarial training to achieve an even stronger robustness.

